\documentclass[conference]{IEEEtran}
\IEEEoverridecommandlockouts
\usepackage{cite}
\usepackage[utf8]{inputenc}
\usepackage{amsmath,amssymb,amsfonts}
\usepackage{algorithmic}
\usepackage{graphicx}
\usepackage{textcomp}
\usepackage{xcolor}
\usepackage{booktabs} % For professional tables
\usepackage{multirow} % For multi-row cells
\usepackage{adjustbox} % For resizing table if needed
\usepackage{subcaption} % More modern package
\usepackage{caption}  
\usepackage{float}

\def\BibTeX{{\rm B\kern-.05em{\sc i\kern-.025em b}\kern-.08em
    T\kern-.1667em\lower.7ex\hbox{E}\kern-.125emX}}
\begin{document}

\title{ADALog: Adaptive Unsupervised Anomaly detection in Logs with Self-attention Masked Language Model\\

}

\author{
    \IEEEauthorblockN{
        Przemek Pospieszny\textsuperscript{*} \quad
        Wojciech Mormul\quad
        Karolina Szyndler\quad
        Sanjeev Kumar
    }
    \\
    \IEEEauthorblockA{
        \textit{NeuroEdge AI, Cognizant}\\
        \{przemyslaw.pospieszny, wojciech.mormul, karolina.szyndler, sanjeev.kumar\}@cognizant.com
    }

}

\iffalse
\author{\IEEEauthorblockN{Przemek Pospieszny\textsuperscript{*} }
\IEEEauthorblockA{\textit{NeuroEdge AI} \\
\textit{Cognizant}\\
przemyslaw.pospieszny@cognizant.com}
\and
\IEEEauthorblockN{Wojciech Mormul}
\IEEEauthorblockA{\textit{NeuroEdge AI} \\
\textit{Cognizant}\\
wojciech.mormul@cognizant.com}
\and
\IEEEauthorblockN{Karolina Szyndler}
\IEEEauthorblockA{\textit{NeuroEdge AI} \\
\textit{Cognizant}\\
karolina.szyndler@cognizant.com}
\and
\IEEEauthorblockN{Sanjeev Kumar}
\IEEEauthorblockA{\textit{NeuroEdge AI} \\
\textit{Cognizant}\\
sanjeev.kumar@cognizant.com}

}
\fi
\maketitle

\begingroup
\renewcommand\thefootnote{\textsuperscript{*}}
\footnotetext{Corresponding author.}
\endgroup

\maketitle

\begin{abstract}
Modern software systems generate extensive heterogeneous log data with dynamic formats, fragmented event sequences, and varying temporal patterns, making anomaly detection both crucial and challenging . To address these complexities, we propose ADALog, an adaptive, unsupervised anomaly detection framework designed for practical applicability across diverse real-world environments. Unlike traditional methods reliant on log parsing, strict sequence dependencies, or labeled data, ADALog operates on individual unstructured logs, extracts intra-log contextual relationships, and performs adaptive thresholding on normal data. The proposed approach utilizes transformer-based, pretrained bidirectional encoder with masked language modeling task,  fine-tuned on normal logs, to capture domain-specific syntactic and semantic patterns essential for accurate anomaly detection in complex environments. Anomalies are identified via token-level reconstruction probabilities, aggregated into log-level scores, with an adaptive percentile-based thresholding, calibrated only on normal data. Through this approach, the model dynamically adapts to evolving system behaviors, contributing to generalization while eliminating the rigidity of heuristic-based thresholds, often utilized in traditional anomaly detection systems. ADALog is systematically evaluated on commonly used log benchmark datasets, BGL, Thunderbird, and Spirit, demonstrating both superior performance and generalization capabilities. While the proposed approach is distinct and not directly comparable, it is outperforming or matching state-of-the-art supervised, self-supervised, and unsupervised log anomaly detection methods, used for reference. We perform comprehensive ablation studies to delve into evaluation of masking strategies, fine-tuning, and token position analysis, impacting model’s decision boundaries to enhance both detection performance and model interpretability. ADALog’s sequence-agnostic, adaptive, and unsupervised approach enables a scalable, resilient, and practical anomaly detection solution, supporting diverse applications across modern software ecosystems, from cloud infrastructures to on-premises computing environments.

\end{abstract}


%%%%%%%%%%%%%
%%%%%%%%%%%%%

\section{Introduction}
In recent decades, software systems have grown increasingly large and complex, spanning from edge applications to massive cloud infrastructures. These environments generate massive stream of heterogeneous log data from devices, IoT sensors, and edge platforms as a log data, which is crucial for understanding software behavior \cite{Candido2021Log-basedStudy}. Logs contain time stamps, error codes, and descriptive messages but due to the large volume of data and the fact that errors do not inherent signal faults. Hence, the only contextual interpretation or hidden patters extraction needs to be extracted, which require complex processing \cite{Yadav2020ALearning}. The logs are of diverse nature and can be either normally sequential or partially or completely unstructured sequences and some systems are able to produce numerical summaries thus compromising the temporal trace continuity. Log parsing tools attempt to impose structure on raw logs but often fail when confronted with diverse or evolving patterns \cite{Wijesinghe2023Log-basedStudy}. Therefore, timely and accurate anomaly detection remains from logs remains challenging but yet crucial for ensuring reliability, stability and risk mitigation of real-time and offline software environments. 

Machine learning-based log anomaly detection is being widely applied for this problem \cite{Landauer2023DeepSurvey}. The methods spans supervised, semi-supervised, and unsupervised with their own advantages and limitations. Supervised and semi-supervised approaches deliver high accuracy with well-labeled datasets but are constrained by the scarcity of labeled anomalies and its low occurrence making their application impractical in real scenarios \cite {Wu2023OnDetection, Ali2023ADetection}. In contrast, self-supervised techniques aim to learn representations from only unlabeled data by solving proxy tasks. However, their effectiveness can be limited by the design of pretext tasks and may not always capture the complex semantics of log data due to noisy pseudo-labels that can lead to suboptimal model performance \cite{YamanakaLogELECTRA:Logs, Guo2021LogBERT:BERT}. Unsupervised methods mitigate all the mentioned constraints by learning normal patterns without presence of labeled anomalies. This makes them well suited for modern and dynamic systems with scarcity of anomalies, novelty of events unseen or captured in prior knowledge \cite{NyyssolaEfficiencyLogs}. Nevertheless, majority of the research and applications in the ML driven log anomaly detection domain depend on parsers that rely on predefined structures leading to potential loss of critical rare information, application of sequence-based models for in in practice often broken sequences of events or anomaly thresholding based on heuristics that reduce generalizability across diverse and dynamic system infrastructures \cite {Khan2024ImpactDetection, Fu2023AnDetection}. The necessity of robust and adaptable frameworks that can properly handle the complexity and variability of the modern log datasets is thus revealed by these limitations.

In order to address these challenges, we propose ADALog, an unsupervised anomaly detection framework that eliminates the need for labeled data, manual parsing, or strict sequence continuity. 
By fine-tuning pretrained transformer-based language model with a self-attention-based Masked Language Model (MLM) on raw log data, ADALog captures domain-specific syntactic and semantic relationships for robust anomaly detection in a variety of log formats. Token-level reconstruction probabilities are aggregated into a log-level score, with an adaptive percentile-based thresholding mechanism replacing heuristic criteria. Comprehensive ablation studies are conducted in respect of masking rates, thresholds and impact of data cleaning, to revel aspects that impact the accuracy and efficiency of the detection. Token position heatmaps are also explored to enhance the interpretability of the approach by identifying the tokens and their positions that differentiate between anomalous and normal log patterns. Finally, a practical approach for deployment in various software environments is presented to bootstrap ADAlog applicability. Our key contributions include:

\begin{enumerate}

    \item We propose an anomaly detection framework that is utilized directly on raw, unparsed system logs and is sequence-agnostic, eliminating the need for complex parsing routines or strict sequential dependencies, thereby enhancing adaptability to diverse logging formats and streamlining deployment.
    \item By leveraging pretrained language model fine-tuned on domain-specific log data through the MLM objective, our approach captures nuanced semantic and syntactic patterns, enabling accurate detection of rare and isolated anomalies in complex and diverse log datasets.
    \item We introduce adaptive thresholding method of normal log prediction errors, enabling robust unsupervised separation of normal and anomalous logs, that adjusts to evolving log behaviors and diverse system architectures without reliance on labeled data.
    \item Through ablation study we explore patterns in log token position influencing model’s decision-making process, enhancing explainability and fostering further research in unsupervised anomaly detection strategies. 
\end{enumerate}

%%%%%%%%%%%%%
%%%%%%%%%%%%%

\section{Related Work}
Log anomaly detection has progressed significantly in the last decades, from heuristic systems towards the application of machine learning techniques covering the entire spectrum of approaches, ranging from supervised and self-supervised to unsupervised learning techniques \cite{Du2017DeepLog:Learning, Yang2021Semi-supervisedEstimation, Zhang2019RobustData, Pan2024RAGLog:Generation, Meng2019Loganomaly:Logs, YamanakaLogELECTRA:Logs, Nedelkoski2020Self-attentiveLogs, Guo2021LogBERT:BERT, Lee2023LAnoBERT:Model, Almodovar2024LogFiT:Models, Markov}. Although these methods have shown a high level of effectiveness when applied to controlled datasets, their practical effectiveness is limited in real-world deployments due to various constraints that occur \cite{Le2022Log-basedWe}. This section critically surveys the modern methods, highlighting their techniques, limitations, and gaps that motivate the proposed ADALog framework.

Initial studies on the application of machine learning in anomaly detection from system logs predominantly followed a supervised paradigm due to effectiveness in utilizing both normal and abnormal patterns for model learning. A notable method such as LogRobust \cite{Zhang2019RobustData} illustrates this strategy by parsing both normal and anomalous logs and feeding sequences of logs into an attention-based Bi-LSTM model to train and extract contextual information and learn different patterns of anomalous events from provided examples. Many of such approaches are extremely accurate when they are trained on extensive and high-quality labeled data and detecting known events. However, reliance on labeled anomaly data introduces scalability challenges in real-world scenarios, where anomalies are rare and time-consuming to annotate \cite{Shah2022AutomatedLearning, Wang2024Log2graphs:Extraction}. Additionally, these methods are limited in their ability to generalize across various unseen log formats, thus being fragile in dynamic environments \cite{Guo2024LogFormer:Detection}.

Self-supervised approaches reduce reliance on labeled data by learning meaningful log representations through proxy tasks. LogBERT \cite{Guo2021LogBERT:BERT} utilizes BERT and a masked language modeling task to detect anomalies within sequences, while LogELECTRA \cite{YamanakaLogELECTRA:Logs} introduces an adversarial token replacement objective, which guides the model to distinguish between real and fake tokens. However, these methods are dependent on log parsing like Drain \cite{Wu2023OnDetection}, which introduces risk of information loss when dealing with unstructured or evolving logs \cite{Fu2023AnDetection}. Furthermore, both methods focus on log context thus limiting usability in environments with broken log sequences \cite{YamanakaLogELECTRA:Logs}. Moreover, LogBERT has been criticized for data leakage during evaluation, where the same logs and patterns are available in train and evaluation dataset, which leads to overoptimistic performance results \cite{Fu2023AnDetection}. 

Unsupervised techniques detect deviations without labeled examples, pseudo-labeled or proxy tasks, by detecting deviations from normal behavior patterns. DeepLog \cite{Du2017DeepLog:Learning} for instance, models logs as sequences using LSTMs in unsupervised manner and identifies normal patterns and anomalous deviations. However, DeepLog’s restriction to sequential modelling and log parsing hinders its applicability in distributed systems with asynchronous logs or unordered sequences. Similarly, LogAnomaly \cite{Meng2019Loganomaly:Logs} extends unsupervised detection by capturing sequential and quantitative anomalies from semantic log embeddings. It is however constrained to parsing based preprocessing and is therefore not as versatile to different log formats. Similarly, TransLog \cite{GuoTRANSLOG:Detection} is dependable on parsers and sequential processing of logs but relies on transformer-based architecture. It does not utilize BERT directly. Instead, it employs a custom Transformer encoder inspired by BERT, incorporating domain-specific log semantics and sequences during pretraining on log datasets. LAnoBERT \cite{Lee2023LAnoBERT:Model} and TransLog have applied transformer architecture to sequential logs. In LAnoBERT, BERT pre-trained model is used as a backbone with MLM tasks to predict reconstruction probabilities. The anomaly detection threshold is defined during the testing phase on the basis of both normal and anomalous data. Additionally, the anomaly score is calculated using only the top k scores for each token in a log, which can result in false positives in practice. TransLog does not utilize BERT directly. Instead, it employs a custom Transformer encoder inspired by BERT, incorporating domain-specific log semantics and sequences during pretraining on log datasets. 

Despite substantial progress, several pervasive limitations continue to impact the broad adoption of existing log anomaly detection methods. First, most approaches are designed to detect abnormalities in sequences of logs, an assumption that is rarely satisfied in practice, where log context is often fragmented or occurs with various temporal frequencies. Next, the parsing-based strategies applied are prone to losing a significant amount of information in case of changes in log structure, resulting in brittle pipelines that fail to handle new or evolving formats. Moreover, heuristic thresholds for anomaly classification are insufficiently flexible to accommodate changing systems and log behaviors. Furthermore, there is no explainability across approaches, especially in self- and unsupervised methods, regarding the nature of the model’s semantic understanding and the features influencing the distinction of anomalies at both the log and token levels. Not to mention data leakage, which is often present and overestimates performance while masking the real usefulness of the model. Collectively, these shortcomings emphasize the need for methods that are robust to fragmented sequences, do not rely on parsing, offer interpretable outputs, avoid rigid heuristics, and remain adaptive in evolving, modern log environments.

%%%%%%%%%%%%%%%%%%%%%%%%%%%
%%%%%%%%%%%%%%%%%%%%%%%%%%%%%%%%%%%%%%%%%%%%%%%%%%%%%%
%%%%%%%%%%%%%%%%%%%%%%%%%%%

\section{Proposed Method - ADALog}
In this section, we present the proposed approach, detailing its key components and methodology. We first define the problem addressed by our framework. This is followed by an explanation of the data preprocessing steps, model training process, and the data postprocessing approach used for anomaly detection.

\begin{figure*}[h]
    \centering
    \includegraphics[width=0.95\textwidth]{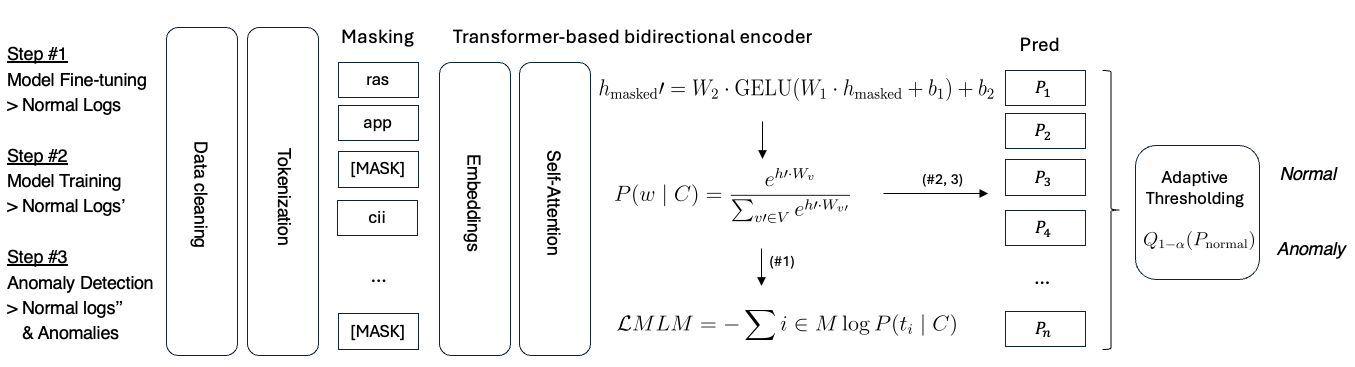} 
    \caption{Overview of proposed ADALog adaptive unsupervised anomaly detection in logs framework.}
    \label{fig:wide}
\end{figure*}

\subsection{Problem Definition}

Let \( l = (w_1, w_2, \dots, w_{\text{slen}}) \) denote a log sequence with \( w_i \in V \), where \( V \) is the set of unique log keys, and let the proposed model \( f \) classify \( l \) as normal or anomalous. During training, only normal log sequences are used, employing a masked language modeling approach \cite{Sinha2021MaskedLittle} where a subset \( M \subset \{1,2,\dots,\text{slen}\} \) of tokens is replaced by a mask token, yielding a masked sequence \(\bar{l}\) such that \(\bar{w}_i = \text{[MASK]}\) for \( i \in M \) and \(\bar{w}_i = w_i\) otherwise. A transformer-based bidirectional encoder \cite{DevlinBERT:Understanding} is then utilized to predict the original token \( w_i \) for each \( i \in M \) by maximizing the conditional probability:

\begin{equation}
\max_{\theta} \sum_{l \in \mathcal{D}_{\text{normal}}} \sum_{i \in M} \log P(w_i \mid C_i; \theta),
\end{equation}

where \( C_i \) represents the contextual tokens surrounding \( w_i \) and \( \theta \) denotes the model parameters. During inference, the model computes an anomaly score \( s(l) \), defined as the negative log-likelihood aggregated over the masked tokens:
\begin{equation}
s(l) = - \frac{1}{|M|} \sum_{i \in M} \log P(w_i \mid C_i; \theta),
\end{equation}
and an adaptive threshold \( T \) is applied such that a log sequence is classified as anomalous if \( s(l) > T \).

\subsection{Preliminaries}\label{AA}

System logs are composed of individual entries that include typical components such as timestamps, log levels, message IDs, file paths, numerical values, and network or memory addresses. Each entry is represented as a variable-length token sequence, reflecting the inherent heterogeneity of system-generated data, and serves as the fundamental unit for our anomaly detection framework.

To standardize raw log data for anomaly detection while preserving adaptability to unseen and evolving formats, we implement a lightweight cleaning procedure that avoids parsing. Given a raw log entry \(l\), which typically comprises components such as timestamps, log levels, message identifiers, file paths, numerical values, and network or memory addresses, the procedure systematically transforms \(l\) into a cleaned representation \(\tilde{l}\). First, any substring conforming to a timestamp pattern (e.g., 2005-06-09-14.53.14.219998) is removed to eliminate temporal bias. Next, compound tokens, particularly those formed by contiguous capital letters, are split to ensure uniform tokenization. Thereafter, file paths are replaced by a generic token filepath to abstract system-specific directory structures, while numerical values (e.g., 3.2143) and network or memory addresses are substituted with the tokens float and address, respectively. This targeted, transformation-based approach minimizes extraneous information and potential bias, yielding a noise-reduced \(\tilde{l}\) that enhances the robustness of the subsequent anomaly detection process.

\subsection{ADALog - Model Approach}

The ADAlog model is built upon the DistilBERT architecture \cite{Sanh2019DistilBERTLighter}, a streamlined variant of BERT, designed to capture intricate contextual relationships while reducing computational overhead. The overall approach is executed in three primary steps, with each step utilizing a distinct subset of normal log data. This separation of datasets in training, threshold selection, and inference is fundamental to maintaining an unsupervised approach, where the model is exposed exclusively to normal logs during training and calibration, thereby eliminating the need for anomaly labels.

\subsubsection{BERT Fine-Tuning on Normal Data}

In this stage, DistilBERT is fine-tuned \cite{Howard2018UniversalClassification} on a dedicated training set of normal log entries: 
\begin{equation}
l = (w_1, w_2, \dots, w_{\text{slen}}),
\end{equation}
drawn from \(\mathcal{D}_{\text{normal}}^{\text{train}}\). For each log entry, 15\% of the tokens are randomly masked; let \( M \subset \{1, 2, \dots, \text{slen}\} \) denote the indices of the masked tokens. The model is trained to predict the original tokens using their surrounding context \( C_i \) \cite{Vaswani2017AttentionNeed}. The prediction probability for a token \( w_i \) is computed via a softmax function over the vocabulary \( V \):
\begin{equation}
P(w_i \mid C_i; \theta) = \frac{\exp(z_{w_i})}{\sum_{w \in V} \exp(z_{w})},
\end{equation}
where \( z_{w} \) is the logit for token \( w \) given the context \( C_i \), and \( \theta \) denotes the model parameters.

The training objective is to minimize the cross-entropy loss for each masked token:
\begin{equation}
\mathcal{L}_{\text{CE}}(w_i, C_i; \theta) = -\log P(w_i \mid C_i; \theta).
\end{equation}
Aggregating this over all masked tokens in the training set yields the overall masked language modeling  loss:
\begin{equation}
\mathcal{L}_{\text{MLM}}(\theta) = 
\sum_{l \in \mathcal{D}_{\text{normal}}^{\text{train}}} 
\sum_{i \in M} \mathcal{L}_{\text{CE}}(w_i, C_i; \theta)
\end{equation}

\begin{equation}
= - \sum_{l \in \mathcal{D}_{\text{normal}}^{\text{train}}} 
\sum_{i \in M} \log P(w_i \mid C_i; \theta).
\end{equation}
Minimization of \(\mathcal{L}_{\text{MLM}}(\theta)\) produces a set of domain-adapted weights \( \theta^* \) that accurately model the statistical distribution, syntax, and semantics of normal log entries.

\subsubsection{Prediction and Threshold Selection}

In the second phase, the fine-tuned model (with weights \( \theta^* \)) is applied to a separate validation set \( \mathcal{D}_{\text{normal}}^{\text{val}} \) to generate token-level predictions and to calibrate an adaptive threshold. To maintain consistency with the training phase, the same masking strategy is applied: for each log entry, 15\% of tokens are randomly masked in batches, indices \( M \) are selected. For each masked token \( w_i \), the model computes the prediction probability:
\begin{equation}
P(w_i \mid C_i; \theta^*) = \frac{\exp(z_{w_i}^*)}{\sum_{w \in V} \exp(z_{w}^*)},
\end{equation}
where \( z_{w}^* \) is the logit computed with the fine-tuned weights \( \theta^* \).

These token-level probabilities are aggregated to yield a log-level anomaly score:
\begin{equation}
s(l) = -\frac{1}{|M|} \sum_{i \in M} \log P(w_i \mid C_i; \theta^*).
\end{equation}
Assuming that the anomaly scores \( s(l) \) for normal logs follow a distribution \( F_S(s) \), the adaptive threshold \( T \) is defined as the 90th percentile (0.9 quantile) of this distribution:
\begin{equation}
T = F_S^{-1}(0.9) \quad \text{such that} \quad 
P(s(l) \leq T) = 0.9, \quad 
\end{equation}
\[
\forall \, l \in \mathcal{D}_{\text{normal}}^{\text{val}}.
\]
Equivalently, if \( f_S(s) \) is the probability density function (PDF) of the anomaly scores, then \( T \) satisfies:
\begin{equation}
\int_{-\infty}^{T} f_S(s) \, ds = 0.9.
\end{equation}
This quantile-based threshold \cite{Tambuwal2021DeepTime-Series} is chosen to balance false positives and recall: setting \( T \) at the 90th percentile ensures that only the top 10\% of normal logs (those with the highest anomaly scores) are near the decision boundary, reducing misclassification of normal logs as anomalies while still capturing true anomalous deviations.

\subsubsection{Anomaly Detection}

In the final phase, the model with weights \( \theta^* \) is deployed on a test dataset \( \mathcal{D}_{\text{test}} \), which comprises both normal and anomalous log entries. For each test log \( l \in \mathcal{D}_{\text{test}} \), the anomaly score is computed using the same masking and aggregation process:
\begin{equation}
s(l) = -\frac{1}{|M|} \sum_{i \in M} \log P(w_i \mid C_i; \theta^*).
\end{equation}
A log entry is classified as anomalous if its score exceeds the adaptive threshold \( T \):
\begin{equation}
\begin{aligned}
    &\text{Log } l \text{ is anomalous if } s(l) > T, \\
    &\text{otherwise, it is classified as normal.}
\end{aligned}
\end{equation}

By employing separate datasets for fine-tuning (\(\mathcal{D}_{\text{normal}}^{\text{train}}\)), threshold selection (\(\mathcal{D}_{\text{normal}}^{\text{val}}\)), and inference (\(\mathcal{D}_{\text{test}}\)), the ADAlog framework adheres to an unsupervised approach. This separation ensures that the model’s calibration is based solely on normal data, making the method applicable in scenarios where anomalous labels are rare or unavailable. Through the integration of cross-entropy loss in the MLM training phase, consistent batch-wise masking during both training and prediction, and rigorous quantile-based threshold selection, ADAlog provides a robust theoretical and practical foundation for anomaly detection in evolving log environments.

%%%%%%%%%%%%%%%%%%%%%%%%%
%%%%%%%%%%%%%%%%%%%%%%%%%
%%%%%%%%%%%%%%%%%%%%%%%%%
%%%%%%%%%%%%%%%%%%%%%%%%%

\section{Experiments}

\begin{table*}[ht]
    \centering
    \caption{Experimental results compared with baseline models on BGL, Thunderbird, and Spirit datasets.}
    \renewcommand{\arraystretch}{1.2} % Adjust row height
    \begin{tabular}{lcccc ccc ccc ccc}
        \toprule
        \multirow{2}{*}{\textbf{Method}} & \multirow{2}{*}{\textbf{Log Parser}} & \multirow{2}{*}{\textbf{Problem}} & \multirow{2}{*}{\textbf{Approach}} & 
        \multicolumn{3}{c}{\textbf{BGL}} & \multicolumn{3}{c}{\textbf{Thunderbird}} & \multicolumn{3}{c}{\textbf{Spirit}} \\
        \cmidrule(lr){5-7} \cmidrule(lr){8-10} \cmidrule(lr){11-13}
        & & & & \textbf{Precision} & \textbf{Recall} & \textbf{F1} & \textbf{Precision} & \textbf{Recall} & \textbf{F1} & \textbf{Precision} & \textbf{Recall} & \textbf{F1} \\
        \midrule
        DeepLog    & Y & Sequential  & Supervised        & 0.90 & 0.83 & 0.86 & 0.90 & 0.90 & 0.90 & 0.02 & 0.99 & 0.04 \\
        LogElectra & Y & Individual  & Self-Supervised   & 0.94 & 0.98 & 0.96 & 0.96 & 0.99 & 0.98 & 0.92 & 0.99 & 0.96 \\
        LogBERT    & Y & Sequential  & Self-Supervised   & 0.89 & 0.92 & 0.90 & 0.96 & 0.96 & 0.96 & 0.92 & 0.98 & 0.95 \\
        Logsy      & N & Sequential  & Self-Supervised   & 0.52 & 0.87 & 0.65 & 0.99 & 1.00 & 0.99 & 0.69 & 0.56 & 0.62 \\
        TransLog   & Y & Sequential  & Unsupervised      & 0.98 & 0.98 & 0.98 & 0.99 & 0.99 & 0.99 & --   & --   & --   \\
        LogAnomaly & N & Sequential  & Unsupervised      & 0.97 & 0.94 & 0.96 & 0.61 & 0.78 & 0.68 & --   & --   & --   \\
        \textbf{ADALog} & \textbf{N} & \textbf{Individual}  & \textbf{Unsupervised} & \textbf{0.91} & \textbf{0.93} & \textbf{0.92} & \textbf{0.90} & \textbf{1.00} & \textbf{0.94} & \textbf{0.92} & \textbf{0.98} & \textbf{0.95} \\
        \bottomrule
    \end{tabular}
\end{table*}

\subsection{Experimental Setup}
The experimental evaluation is performed on three widely recognized datasets BGL, Thunderbird and Spirit \cite{Oliner2007WhatLogs}. The BGL (BlueGene/L) dataset is captured from a production system of a large computer cluster and contains system event and hardware logs. Thunderbird and Spirit are both derived from high-performance computing clusters and include system, application, job, mode and security log events. They differ in terms for log volume, granularity, patterns, event frequency, and foremost nature of anomalous events and their isolation towards sequence of events. Prior to model training, the raw logs from each dataset are cleaned with approach described in section 3, which involves formatting the logs to remove unnecessary information such as time stamps, splitting compound tokens, and replacing file paths, numerical values and addresses with generic placeholders. After the cleaning, the BGL dataset had 211,504 logs, 16,099 unique normal logs and 97 anomalies, the Thunderbird dataset had 278,960 logs, 35,485 unique normal logs and 1,019 anomalies, and the Spirit dataset had 524,473 logs, 5,622 unique normal logs and 55 anomalies.

To ensure the integrity of unsupervised model training and evaluation, the experimental data was randomly divided (70/15/15) into unique sets of distinct logs for each stage of the model training and evaluation, as described in Section 3. Specifically, 70\% of unique normal logs were utilized for the fine-tuning process of the DistilBERT pre-trained model for Masked Language Model tasks, allowing the model to learn the domain-specific patterns, syntax, and semantics inherent in normal log token data without any contamination from anomalous entries. 

As a backbone, distilbert-base-uncased \cite{Sanh2019DistilBERTLighter} is employed and trained for 10 epochs with a batch size of 64. In the fine-tuning stage, the cross-entropy loss is minimized to update the model parameters so that the model can capture the typical patterns inherent in normal logs. The remaining 15\% of the unique normal logs are held for the prediction phase and threshold determination, based on the distribution of scores for normal logs and the 0.9 quantile selection. During the prediction phase, the fine-tuned model is applied to these normal logs using the same 15\% random masking strategy employed in the previous step of model training. Next, for each log, an anomaly score is computed as the negative average log-probability on the masked tokens, reflecting how well the log conforms to the learned normal patterns.

Threshold selection is performed exclusively on the set of normal logs allocated for prediction. The threshold is defined as the 90th percentile of the anomaly score distribution derived from these logs, ensuring that the threshold is determined entirely by normal behavior. This design choice ensures that only the most atypical 10\% of the normal logs, those with the highest anomaly scores—are near the decision boundary, thereby balancing the tradeoff between false positives and recall. Finally, during the inference phase, the model is evaluated on a test set that contains both normal logs, 15\% of the unique normal logs, kept separate from the training and validation sets, and all available anomalous logs. Each test log is classified as anomalous if its computed anomaly score exceeds the predefined threshold.

Although our approach is distinct in that it does not rely on log parsing and operates on individual logs in an unsupervised manner, we provide comparative performance benchmarks against several state-of-the-art methods, including DeepLog \cite{Du2017DeepLog:Learning}, LogElectra \cite{YamanakaLogELECTRA:Logs}, LogBERT \cite{Guo2021LogBERT:BERT}, Logsy \cite{Nedelkoski2020Self-attentiveLogs}, TransLog \cite{GuoTRANSLOG:Detection}, and LogAnomaly \cite{Meng2019Loganomaly:Logs}. Evaluation is carried out using standard metrics such as Precision, Recall, and F1 score to facilitate a comprehensive and rigorous comparison of anomaly detection performance. It needs to be noted that Performance of benchmark models are reported from Logsy and LogBert and TransLog research papers.

\subsection{Results}
We present the main experimental results of our approach compared to baseline methods on the BGL, Thunderbird, and Spirit datasets. We also conduct ablation studies on understand influencing factors on overall model performance, in particular token batch size for MLM and impact of fine-tuning of log data. Finally, we examined impact of token position for normal and anomaly log detection to contribute to explainability of model’s decision boundaries.

\subsubsection{ADALog performance}
Table 1 compares the performance of ADAlog and baseline methods, including DeepLog, LogElectra, LogBERT, Logsy, TransLog, and LogAnomaly on the BGL, Thunderbird and Spirit datasets. Although these baselines serve as valuable references, they are not directly comparable to ADAlog because of significant methodological differences, such as log parsing, the availability of labeled anomalies, and the modeling of sequential log dependencies. Nevertheless, they serve as broad benchmarks for contextualizing typical performance ranges in anomaly detection on these datasets.

ADALog performs consistently well, achieving F1 scores of 0.92, 0.94 and 0.95 on the BGL, Thunderbird, and Spirit datasets respectively. This level of performance is particularly notable given the divergent characteristics of each dataset. BGL logs, which are derived from a high performance computing cluster, are dense and have repetitive patterns that are indicative of highly optimized system processes. In contrast, the Thunderbird dataset, which is a large scale distributed system's logs, is known to have more event type and message structural heterogeneity, which makes the task of identifying rare anomalies challenging for any model. On the other hand, the Spirit dataset which is also derived from another high-performance computing system has a intermediate level of formatting complexity and error rates to test the generality of the model.

In this diverse setting, ADALog maintains a robust balance of precision and recall. For instance, on Thunderbird, it achieves a recall of 1.00 at a precision of 0.90, which means that it is very sensitive to anomalous events without having a high false positive rate. Such strong recall rate in a diverse log environment suggests that ADALog is able to learn the statistical distribution of normal logs well even when the logs are of different types and may contain mixed formats or overlapping tokens. On the BGL dataset, the model achieves a precision of 0.91 and a recall of 0.93, which means that it is able to identify the majority of the true anomalies without labeling many normal instances as anomalous. This balanced performance is particularly valuable in operational contexts where false positives can lead to unnecessary investigative overhead, yet missed anomalies can carry severe consequences, such as system downtime or security breaches.

Furthermore, each of these datasets exhibits unique error signatures, distribution of log events, and practical challenges in anomaly identification. The persistent strength of ADALog across all three points to a capacity for learning generalizable representations of normal log behavior, rather than overfitting to idiosyncratic token patterns or superficial data traits. In scenarios where log formats evolve rapidly or vary across different subsystems, the ability to recognize anomalies based on core contextual cues, rather than reliance on fixed templates or labeled examples, provides a substantial advantage.

\begin{figure*}[t]
    \centering
    \begin{subfigure}[b]{0.3\linewidth}
        \centering
        \includegraphics[height=4cm]{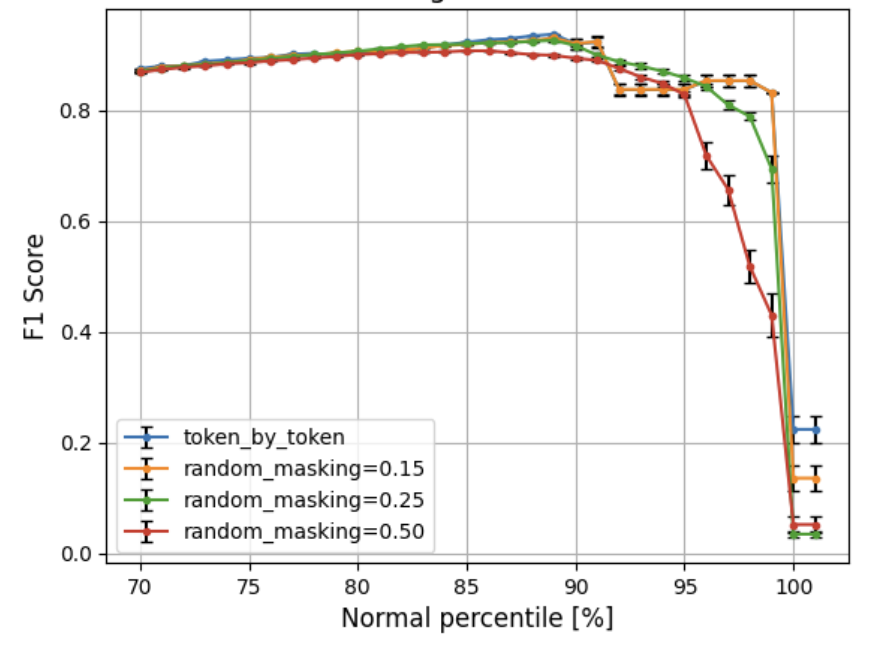}
        \caption{BGL}
        \label{fig:sub1}
    \end{subfigure}
    \hfill
    \begin{subfigure}[b]{0.3\linewidth}
        \centering
        \includegraphics[height=4cm]{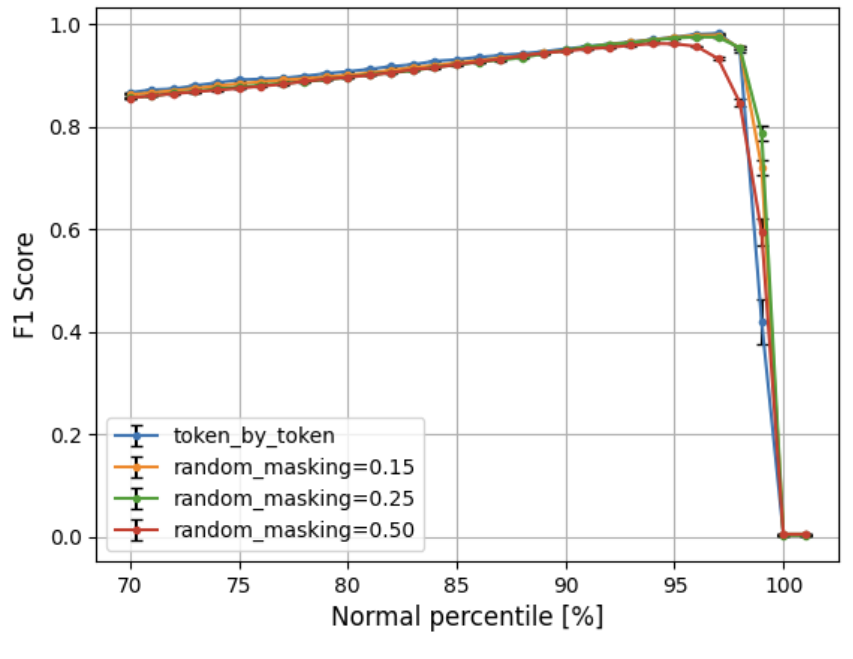}
        \caption{Thunderbird}
        \label{fig:sub2}
    \end{subfigure}
    \hfill
    \begin{subfigure}[b]{0.3\linewidth}
        \centering
        \includegraphics[height=4cm]{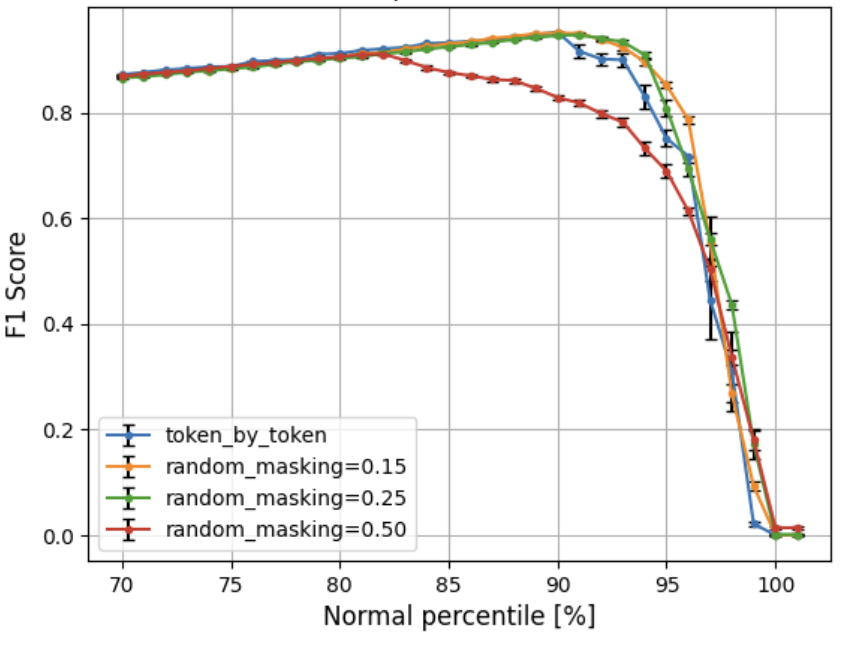}
        \caption{Spirit}
        \label{fig:sub3}
    \end{subfigure}

    \caption{Impact of different token masking strategies on anomaly detection performance across three benchmark log datasets.}
    \label{fig:three_images_twocolumn}
\end{figure*}

\subsubsection{Ablation studies}
In the first part of our ablation study, we examined the effect of varying the fraction of masked tokens during prediction and inference in the MLM objective, prior aggregation and thresholding. Specifically, we compared four masking strategies, such as masking every token individually, masking a random 15\% batch of tokens, as used in our main results, masking 25\%, and masking 50\%. As depicted in Figure 2, there is a noticeable difference in F1 scores across applied strategies presented over a range of normal logs percentile-based thresholds from 70\% to 100\%. In principle, the 15\% masking consistently yields either the best or near-best performance across all three datasets, indicating that it provides a strong balance between exposing enough context to the model and forcing the model to learn robust contextual representations of normal log patterns.

One of the most notable observations is that masking a larger proportion of tokens, such as 25\% or 50\%, can lead to a steeper drop in F1 scores beyond the 90th–95th percentile thresholds, suggesting higher variance in the anomaly scores and more sensitivity to minor deviations in normal log patterns. The token-by-token strategy, in contrast, tends to generate similar results or marginally underperform when compared to random 15\% and 25\% batch masking. This suggests that masking every token individually, although intuitively should yield the best results, may, to some degree, reduce the model’s ability to learn nuanced patterns if compared to batch masking, where multiple random tokens are hidden, and the model is better encouraged to extract natural contextual relationships among tokens, leading to more effective learning. Furthermore, it needs to be noted that the reduced model cycles to predict masked tokens are less computationally demanding and significantly improve the inference time. Therefore, the random masking of 15\% of tokens appears to strike the optimal balance, since the model is adequately challenged during training to learn log sequence representation, and inference time is not overly extended. 

The above supports our choice of applied masking strategy in our main experiments. The presented results underline that a moderate masking rate can elicit deeper contextual learning and more discriminative anomaly scores, whereas extreme masking rates or a token-by-token approach may compromise the fine-grained understanding of normal log behavior that is crucial for reliable anomaly detection.

\iffalse
\begin{figure}[h]
    \centering
    \begin{subfigure}[b]{0.3\textwidth}
        \centering
        \includegraphics[width=\textwidth]{BGL_masking.png}
        \caption{lorem ipsum}
        \label{fig:sub1}
    \end{subfigure}
    \hfill
    \begin{subfigure}[b]{0.3\textwidth}
        \centering
        \includegraphics[width=\textwidth]{Thunder_masking.png}
        \caption{lorem ipsum}
        \label{fig:sub2}
    \end{subfigure}
    \hfill
    \begin{subfigure}[b]{0.3\textwidth}
        \centering
        \includegraphics[width=\textwidth]{Spirit_masking.png}
        \caption{lorem ipsum}
        \label{fig:sub3}
    \end{subfigure}
    
    \caption{lorem ipsum}
    \label{fig:three_images_row}
\end{figure}
\fi

An important aspect of our experiments was the direct comparison of a pretrained language model before and after fine-tuning on normal log data, which revealed that the impact of fine-tuning is extremely significant. Without domain-specific fine-tuning, the model’s outputs were nearly random, as it failed to capture the subtle syntactic and semantic patterns inherent in system logs. Evaluations on benchmarking BGL, Thunderbird and Spirit datasets indicated that while the pretrained model’s performance plateaued at low accuracy levels, its anomaly detection capability remaining largely insensitive to variations in the anomaly threshold, and the fine-tuned model consistently maintained superior accuracy over the entire range of thresholds. This marked performance gap illustrates that general text pretraining does not suffice for the unique demands of log analysis since system logs often contain specialized tokens, hardware or software specific message formats, and recurring patterns of normal behavior that are absent in typical pretraining corpora. By fine-tuning on normal log data, the model learns log specific regularities that can range from numeric placeholders, address tokens to recurrent subsystem messages. Through this it can reduce false positives and significantly enhancing recall. In essence, this analysis underscores importance of domain adaptation in transforming a generic pretrained model into a robust and accurate tool for anomaly detection in complex software environments. 

\begin{figure}[h]
    \centering
    \begin{minipage}{0.49\linewidth}
        \centering
        \subfloat[BGL Normal Log]{%
            \includegraphics[width=\linewidth]{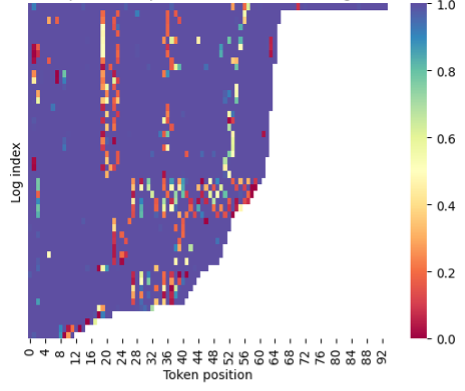}
        }
        \vspace{0.5em}
        \subfloat[Thunderbird Normal Log]{%
            \includegraphics[width=\linewidth]{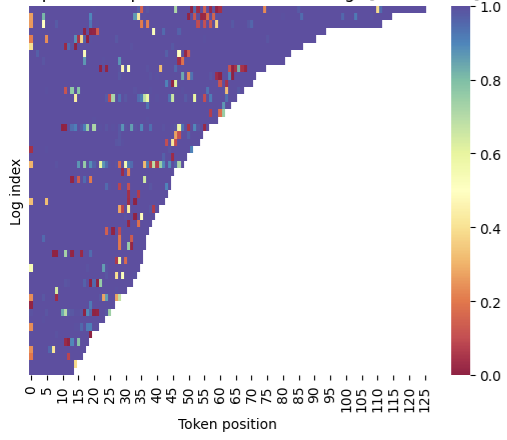}
        }
    \end{minipage}
    \hfill
    \begin{minipage}{0.49\linewidth}
        \centering
        \subfloat[BGL Anomaly]{%
            \includegraphics[width=\linewidth]{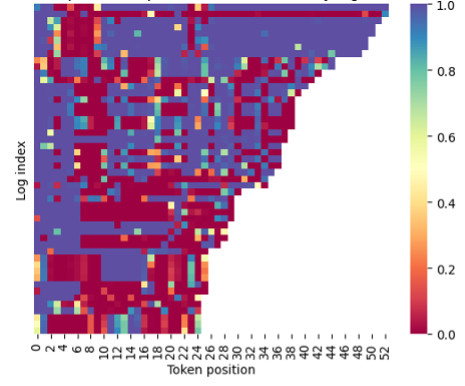}
        }
        \vspace{0.5em}
        \subfloat[Thunderbird Anomaly]{%
            \includegraphics[width=\linewidth]{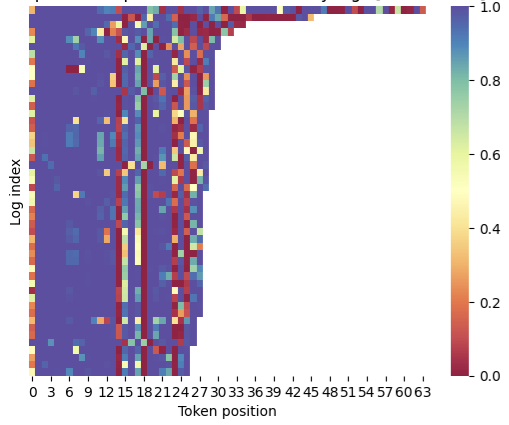}
        }
    \end{minipage}
    
    \vspace{1em}
    
    \begin{minipage}{0.49\linewidth}
        \centering
        \subfloat[Spirit Normal Log]{%
            \includegraphics[width=\linewidth]{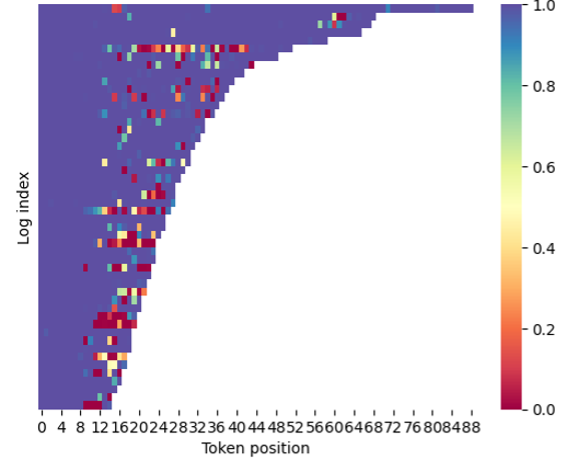}
        }
    \end{minipage}
    \hfill
    \begin{minipage}{0.49\linewidth}
        \centering
        \subfloat[Spirit Anomaly]{%
            \includegraphics[width=\linewidth]{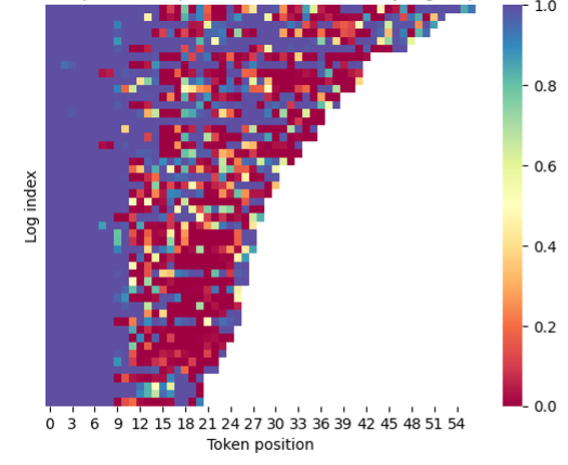}
        }
    \end{minipage}
    
    \caption{Heatmap of token probabilities for BGL, Thunderbird, and Spirit logs}
    \label{fig:combined}
\end{figure}

To enhance the explainability of ADAlog’s anomaly detection capabilities, we performed an additional ablation study to determine the influence of token position on the MLM prediction probabilities of normal and anomalous logs from the BGL, Thunderbird, and Spirit datasets. Figure 3 depict the heatmaps of these probabilities, with the x-axis presenting the token positions within log entries, the y-axis the log indices, and the color intensity outlining the model’s confidence in predicting the masked tokens, from low probability in red to high probability represented in blue. A clear and consistent trend emerges across all datasets, normal logs exhibit higher prediction probabilities concentrated in contiguous regions, particularly in the middle-to-late token positions. This is especially evident in the BGL and Thunderbird datasets, where the heatmaps for normal logs have dominant blue regions with very few interruptions, which indicates that the model is very confident in predicting tokens that are in the sequence of learned syntactic and semantic rules. In contrast, anomalous logs display fragmented and irregular patterns, characterized by scattered low-probability regions dispersed throughout the token positions. This disruption in the model’s confidence when presented with unexpected token sequences is a hallmark of anomalous behavior.

The system logs are essentially semi-structured, with structural variations across the BGL, Thunderbird, and Spirit datasets that affect the probability distribution of token predictions. The BGL logs, which are generated from high-performance computing environments, have well-organized system events and rather homogeneous prefixes, such as hardware status symbols and system notes. The initial tokens in these logs tend to be highly predictable, resulting in consistently high prediction probabilities at the start of log entries, even in anomalous cases. In contrast, central and latter tokens hold the values of various system parameters, error messages, and the performance metrics of the running system, and anomalies are seen as deviations from normal behavior, which is reflected by a sharp decrease in the prediction probabilities. Thunderbird logs, which are responsible for describing activities in distributed systems, are more diverse in their nature due to the inclusion of network-related events, process identifiers, and system calls whose output can vary depending on the specific situation. This increases the complexity, which is shown in the heatmaps, where the confidence in the prediction is reduced dramatically after the 20–30th token positions, which correspond to volatile fields like pointers and variables. Spirit logs, that monitor system failures and filesystem transactions, like the previous sets of logs, have high prediction probabilities for static metadata at the beginning of the log entries, with anomalies occurring in the areas that contain a large number of dynamic variables such as file paths and session identifiers.

These observations reinforce the notion that token position contextual dynamics are crucial to ADAlog’s anomaly detection capabilities. The heatmaps demonstrate that anomalies are often associated with localized disruptions in token prediction probabilities within individual logs, particularly in the middle-to-late token positions where contextual dependencies are strongest. Instead of analyzing sequences of logs, ADAlog focuses on capturing the internal structure of each log entry, modeling token relationships within a single log. The model’s sensitivity to disruptions in these intra-log contextual relationships, unexpected token sequences, irregular parameter values, or syntactic pattern deviations allows the detection of abnormal events. The results of the scattered low-probability regions in anomalous logs suggest that even subtle inconsistencies in token dependencies can trigger correct anomaly detection. Furthermore, the concentration of the prediction confidence shift in particular token positions, as derived from the heatmaps, directly reveals which parts of the log the model finds most informative in distinguishing between normal and anomalous behaviors. Certainly, the probability distributions and token position analysis enhance our understanding of the model’s decision-making process and contribute to the explainability of its anomaly detection capabilities, which could be applied for further work.
\iffalse
 \begin{figure}[h]
    \centering
    \subfloat[Normal Log]{%
        \includegraphics[width=0.45\linewidth]{BGL_Normal.png}
    }
    \hfill
    \subfloat[Anomaly]{%
        \includegraphics[width=0.45\linewidth]{BGL_Anomalies.png}
    }
    \caption{Heatmap of token probabilities for BGL logs}
    \label{fig:side_by_side}
\end{figure}

\begin{figure}[h]
    \centering
    \subfloat[Normal Log]{%
        \includegraphics[width=0.45\linewidth]{Thunder Normal.png}
    }
    \hfill
    \subfloat[Anomaly]{%
        \includegraphics[width=0.45\linewidth]{Thunder Anomalies.png}
    }
    \caption{Heatmap of token probabilities for Thunderbird logs}
    \label{fig:side_by_side}
\end{figure}

\begin{figure}[h]
    \centering
    \subfloat[Normal Log]{%
        \includegraphics[width=0.45\linewidth]{Spirit Normal.png}
    }
    \hfill
    \subfloat[Anomaly]{%
        \includegraphics[width=0.45\linewidth]{Spirit Anomalies.png}
    }
    \caption{Heatmap of token probabilities for Spirit logs}
    \label{fig:side_by_side}
\end{figure}
\fi
%%%%%%%

%%%%%%%%%%%%%%%%%%%%%%%%%
%%%%%%%%%%%%%%%%%%%%%%%%%
%%%%%%%%%%%%%%%%%%%%%%%%%
%%%%%%%%%%%%%%%%%%%%%%%%%
\section{Discussion and Limitations}
We propose an unsupervised transformer-based mask language model for detecting anomalies in system logs demonstrating strong performance and generalization capabilities. By operating on individual log entries and learning from exclusively normal data through a masked language modeling objective, the method is inherently adaptable to evolving log formats and heterogeneous log structures. The unsupervised framework of this method enables it to detect anomalous events by the deviations from the learned normal behavior which makes it suitable for wide range of use cases, including automotive predictive maintenance, medical device monitoring, IoT device behavior detection. 

In practical terms, the methodology, like other transformer-based log anomaly detection approaches, is most appropriate for offline analysis of log data in batches and when there are computational resources available. The model’s reliance on language model architectures, such as DistilBERT, ensures high accuracy in capturing the subtle syntactic and semantic patterns of normal logs, leading to effective anomaly detection. However, this advantage is accompanied by a high cost of computational resources during training and inference. Therefore, the current implementation may not be readily applicable for real-time anomaly detection in low-latency environments, such as near-edge or far-edge deployments in resource-constrained environments.

To overcome these limitations and improve the real-time performance on edge devices, the future work should be directed towards the application of the techniques such as knowledge distillation and knowledge transfer. A promising strategy is to deploy a hybrid architecture where a large unsupervised model operates in the cloud to continuously process unseen log data. The concept relies on a principle that the cloud-based model learns robust representations through unsupervised learning and can periodically transfer its knowledge to a smaller, lightweight supervised model designed to run on edge devices. In this system, edge would transmit data embeddings or summaries to the cloud, allowing the large model to update its internal representations continuously. In turn, these updated representations would be employed to update the smaller model and thus maintain its consistency with the changing nature of the data. This way, the hybrid model achieves the accuracy and robustness of transformer-based unsupervised models combined with the timeliness and computational budget of the model that is necessary for real-time anomaly detection on edge devices.

Future work should also consider evaluating the applicability and efficiency of the proposed method across a wide range of practical use cases. This includes domains spanning from healthcare to manufacturing and cover such use cases like IoT device monitoring, real-time anomaly detection in automotive systems or even financial transaction monitoring. Through such evaluation the practicality and generalization of proposed approach could be assessed under varying real word operational systems, events and data. Such evaluations will not only strengthen the validity of the method but also reveal the opportunities for improvement for specific application areas, including for offline event detection and real-time and resource-limited applications.

\section{Conclusion}
In this paper, we introduce ADALog, an innovative unsupervised anomaly detection framework that leverages transformer-based masked language modeling and adaptive thresholding to extract anomalies without the need for strict sequence dependencies. ADALog reflects the requirements of current dynamic software environments, characterized by changing log formats and events, heterogeneous data structures, and complex system interactions. The proposed method effectively captures intricate syntactic and semantic patterns across diverse individual log formats and token sequences due to the utilized approach of MLM-pretrained DistilBERT fine-tuning, token prediction-based token scoring, and percentile-based thresholding, all performed on normal log behavior. Our experimental evaluation on commonly utilized benchmark datasets demonstrates ADALog’s ability to generalize, achieving a robust balance between precision and recall. The application of the transformer architecture empowers the model’s deep understanding of contextual relationships at the token level, ultimately enhancing both detection accuracy and interpretability, as presented in the comprehensive ablation study.

The subsequent studies should foremost concentrate on expanding ADALog’s application for low-latency edge deployment via a hybrid cloud-edge approach, which would seamlessly integrate the high accuracy of transformer-based models with the efficiency required for real-time processing in diverse edge-driven operational environments.

\bibliographystyle{IEEEtran} 
\bibliography{Anomaly_Log_Paper}

\begin{thebibliography}{10}
\providecommand{\url}[1]{#1}
\csname url@samestyle\endcsname
\providecommand{\newblock}{\relax}
\providecommand{\bibinfo}[2]{#2}
\providecommand{\BIBentrySTDinterwordspacing}{\spaceskip=0pt\relax}
\providecommand{\BIBentryALTinterwordstretchfactor}{4}
\providecommand{\BIBentryALTinterwordspacing}{\spaceskip=\fontdimen2\font plus
\BIBentryALTinterwordstretchfactor\fontdimen3\font minus \fontdimen4\font\relax}
\providecommand{\BIBforeignlanguage}[2]{{%
\expandafter\ifx\csname l@#1\endcsname\relax
\typeout{** WARNING: IEEEtran.bst: No hyphenation pattern has been}%
\typeout{** loaded for the language `#1'. Using the pattern for}%
\typeout{** the default language instead.}%
\else
\language=\csname l@#1\endcsname
\fi
#2}}
\providecommand{\BIBdecl}{\relax}
\BIBdecl

\bibitem{Candido2021Log-basedStudy}
J.~C{\^{a}}ndido, M.~Aniche, and A.~V. Deursen, ``{Log-based software monitoring: a systematic mapping study},'' \emph{PeerJ Computer Science}, vol.~7, 2021.

\bibitem{Yadav2020ALearning}
R.~B. Yadav, P.~S. Kumar, and S.~V. Dhavale, ``{A Survey on Log Anomaly Detection using Deep Learning},'' in \emph{ICRITO 2020 - IEEE 8th International Conference on Reliability, Infocom Technologies and Optimization (Trends and Future Directions)}, 2020.

\bibitem{Wijesinghe2023Log-basedStudy}
N.~Wijesinghe and H.~Hemmati, ``{Log-based Anomaly Detection of Enterprise Software: An Empirical Study},'' in \emph{IEEE International Conference on Software Quality, Reliability and Security, QRS}, 2023.

\bibitem{Landauer2023DeepSurvey}
M.~Landauer, S.~Onder, F.~Skopik, and M.~Wurzenberger, ``{Deep learning for anomaly detection in log data: A survey},'' \emph{Machine Learning with Applications}, vol.~12, 2023.

\bibitem{Wu2023OnDetection}
X.~Wu, H.~Li, and F.~Khomh, ``{On the effectiveness of log representation for log-based anomaly detection},'' \emph{Empirical Software Engineering}, vol.~28, no.~6, 2023.

\bibitem{Ali2023ADetection}
\BIBentryALTinterwordspacing
S.~Ali, C.~Boufaied, D.~Bianculli, P.~Branco, and L.~Briand, ``{A Comprehensive Study of Machine Learning Techniques for Log-Based Anomaly Detection},'' 7 2023. [Online]. Available: \url{https://arxiv.org/abs/2307.16714v3}
\BIBentrySTDinterwordspacing

\bibitem{YamanakaLogELECTRA:Logs}
Y.~Yamanaka, T.~Takahashi, T.~Minami, and Y.~Nakajima, ``{LogELECTRA: Self-supervised Anomaly Detection for Unstructured Logs}.''

\bibitem{Guo2021LogBERT:BERT}
H.~Guo, S.~Yuan, and X.~Wu, ``{LogBERT: Log Anomaly Detection via BERT},'' in \emph{Proceedings of the International Joint Conference on Neural Networks}, vol. 2021-July, 2021.

\bibitem{NyyssolaEfficiencyLogs}
\BIBentryALTinterwordspacing
J.~Nyyss{\"{o}}l{\"{a}} and M.~M{\"{a}}ntyl{\"{a}}, ``{Efficiency of Unsupervised Anomaly Detection Methods on Software Logs},'' \emph{Proceedings of ACM Conference (Conference'17)}, vol.~1. [Online]. Available: \url{https://github.com/logpai/loglizer}
\BIBentrySTDinterwordspacing

\bibitem{Khan2024ImpactDetection}
\BIBentryALTinterwordspacing
Z.~A. Khan, D.~Shin, D.~Bianculli, and L.~C. Briand, ``{Impact of log parsing on deep learning-based anomaly detection},'' \emph{Empirical Software Engineering}, vol.~29, no.~6, pp. 1--33, 11 2024. [Online]. Available: \url{https://link.springer.com/article/10.1007/s10664-024-10533-w}
\BIBentrySTDinterwordspacing

\bibitem{Fu2023AnDetection}
Y.~Fu, M.~Yan, Z.~Xu, X.~Xia, X.~Zhang, and D.~Yang, ``{An empirical study of the impact of log parsers on the performance of log-based anomaly detection},'' \emph{Empirical Software Engineering}, vol.~28, no.~1, 2023.

\bibitem{Du2017DeepLog:Learning}
M.~Du, F.~Li, G.~Zheng, and V.~Srikumar, ``{DeepLog: Anomaly detection and diagnosis from system logs through deep learning},'' in \emph{Proceedings of the ACM Conference on Computer and Communications Security}, 2017.

\bibitem{Yang2021Semi-supervisedEstimation}
L.~Yang, J.~Chen, Z.~Wang, W.~Wang, J.~Jiang, X.~Dong, and W.~Zhang, ``{Semi-supervised log-based anomaly detection via probabilistic label estimation},'' in \emph{Proceedings - International Conference on Software Engineering}, 2021.

\bibitem{Zhang2019RobustData}
X.~Zhang, Y.~Xu, Q.~Lin, B.~Qiao, H.~Zhang, Y.~Dang, C.~Xie, X.~Yang, Q.~Cheng, Z.~Li, J.~Chen, X.~He, R.~Yao, J.~G. Lou, M.~Chintalapati, F.~Shen, and D.~Zhang, ``{Robust log-based anomaly detection on unstable log data},'' in \emph{ESEC/FSE 2019 - Proceedings of the 2019 27th ACM Joint Meeting European Software Engineering Conference and Symposium on the Foundations of Software Engineering}, 2019.

\bibitem{Pan2024RAGLog:Generation}
J.~Pan, W.~S. Liang, and Y.~Yidi, ``{RAGLog: Log Anomaly Detection using Retrieval Augmented Generation},'' in \emph{Proceedings - 2024 IEEE World Forum on Public Safety Technology, WFPST 2024}, 2024.

\bibitem{Meng2019Loganomaly:Logs}
W.~Meng, Y.~Liu, Y.~Zhu, S.~Zhang, D.~Pei, Y.~Liu, Y.~Chen, R.~Zhang, S.~Tao, P.~Sun, and R.~Zhou, ``{Loganomaly: Unsupervised detection of sequential and quantitative anomalies in unstructured logs},'' in \emph{IJCAI International Joint Conference on Artificial Intelligence}, vol. 2019-August, 2019.

\bibitem{Nedelkoski2020Self-attentiveLogs}
S.~Nedelkoski, J.~Bogatinovski, A.~Acker, J.~Cardoso, and O.~Kao, ``{Self-attentive classification-based anomaly detection in unstructured logs},'' in \emph{Proceedings - IEEE International Conference on Data Mining, ICDM}, vol. 2020-November, 2020.

\bibitem{Lee2023LAnoBERT:Model}
Y.~Lee, J.~Kim, and P.~Kang, ``{LAnoBERT: System log anomaly detection based on BERT masked language model},'' \emph{Applied Soft Computing}, vol. 146, 2023.

\bibitem{Almodovar2024LogFiT:Models}
C.~Almodovar, F.~Sabrina, S.~Karimi, and S.~Azad, ``{LogFiT: Log Anomaly Detection Using Fine-Tuned Language Models},'' \emph{IEEE Transactions on Network and Service Management}, vol.~21, no.~2, 2024.

\bibitem{Markov}
R.~Mojarad and H.~Zarandi, ``Markov-based anomaly correction in embedded systems,'' \emph{International Journal of Computer Theory and Engineering}, vol.~8, pp. 272--279, 08 2016.

\bibitem{Le2022Log-basedWe}
V.~H. Le and H.~Zhang, ``{Log-based Anomaly Detection with Deep Learning: How Far Are We?}'' in \emph{Proceedings - International Conference on Software Engineering}, vol. 2022-May, 2022.

\bibitem{Shah2022AutomatedLearning}
A.~H. Shah, D.~Pasha, E.~H. Zadeh, and S.~Konur, ``{Automated Log Analysis and Anomaly Detection Using Machine Learning},'' \emph{Frontiers in Artificial Intelligence and Applications}, vol. 358, pp. 137--147, 10 2022.

\bibitem{Wang2024Log2graphs:Extraction}
\BIBentryALTinterwordspacing
C.~Wang, D.~Xu, and Z.~Li, ``{Log2graphs: An Unsupervised Framework for Log Anomaly Detection with Efficient Feature Extraction},'' 9 2024. [Online]. Available: \url{http://arxiv.org/abs/2409.11890}
\BIBentrySTDinterwordspacing

\bibitem{Guo2024LogFormer:Detection}
\BIBentryALTinterwordspacing
H.~Guo, J.~Yang, J.~Liu, J.~Bai, B.~Wang, Z.~Li, T.~Zheng, B.~Zhang, J.~Peng, and Q.~Tian, ``{LogFormer: A Pre-train and Tuning Pipeline for Log Anomaly Detection},'' \emph{Proceedings of the AAAI Conference on Artificial Intelligence}, vol.~38, no.~1, pp. 135--143, 1 2024. [Online]. Available: \url{https://arxiv.org/abs/2401.04749v1}
\BIBentrySTDinterwordspacing

\bibitem{GuoTRANSLOG:Detection}
H.~Guo, X.~Lin, J.~Yang, Y.~Zhuang, J.~Bai, T.~Zheng, L.~Zheng, W.~Hou, B.~Zhang, and Z.~Li, ``{TRANSLOG: A Unified Transformer-based Framework for Log Anomaly Detection}.''

\bibitem{Sinha2021MaskedLittle}
K.~Sinha, R.~Jia, D.~Hupkes, J.~Pineau, A.~Williams, and D.~Kiela, ``{Masked Language Modeling and the Distributional Hypothesis: Order Word Matters Pre-training for Little},'' in \emph{EMNLP 2021 - 2021 Conference on Empirical Methods in Natural Language Processing, Proceedings}, 2021.

\bibitem{DevlinBERT:Understanding}
\BIBentryALTinterwordspacing
J.~Devlin, M.-W. Chang, K.~Lee, K.~T. Google, and A.~I. Language, ``{BERT: Pre-training of Deep Bidirectional Transformers for Language Understanding}.'' [Online]. Available: \url{https://github.com/tensorflow/tensor2tensor}
\BIBentrySTDinterwordspacing

\bibitem{Sanh2019DistilBERTLighter}
\BIBentryALTinterwordspacing
V.~Sanh, L.~Debut, J.~Chaumond, and T.~Wolf, ``{DistilBERT, a distilled version of BERT: smaller, faster, cheaper and lighter},'' 10 2019. [Online]. Available: \url{https://arxiv.org/abs/1910.01108v4}
\BIBentrySTDinterwordspacing

\bibitem{Howard2018UniversalClassification}
J.~Howard and S.~Ruder, ``{Universal language model fine-tuning for text classification},'' in \emph{ACL 2018 - 56th Annual Meeting of the Association for Computational Linguistics, Proceedings of the Conference (Long Papers)}, vol.~1, 2018.

\bibitem{Vaswani2017AttentionNeed}
A.~Vaswani, N.~Shazeer, N.~Parmar, J.~Uszkoreit, L.~Jones, A.~N. Gomez, L.~Kaiser, and I.~Polosukhin, ``{Attention is all you need},'' in \emph{Advances in Neural Information Processing Systems}, vol. 2017-December, 2017.

\bibitem{Tambuwal2021DeepTime-Series}
A.~I. Tambuwal and D.~Neagu, ``{Deep Quantile Regression for Unsupervised Anomaly Detection in Time-Series},'' \emph{SN Computer Science}, vol.~2, no.~6, 2021.

\bibitem{Oliner2007WhatLogs}
A.~Oliner and J.~Stearley, ``{What supercomputers say: A study of five system logs},'' in \emph{Proceedings of the International Conference on Dependable Systems and Networks}, 2007.

\end{thebibliography}

\end{document}